\documentclass{article}

\PassOptionsToPackage{numbers}{natbib}

\usepackage[preprint]{nips_2018}

\usepackage[utf8]{inputenc} 
\usepackage[T1]{fontenc}    
\usepackage{url}            
\usepackage{booktabs}       
\usepackage{amsfonts}       
\usepackage{nicefrac}       
\usepackage{microtype}      

\usepackage[colorlinks=true,linkcolor=black,citecolor=black,urlcolor=black]{hyperref}
\usepackage{color}
\usepackage{amsmath}
\usepackage{amssymb}
\usepackage{graphicx}
\usepackage[outdir=./]{epstopdf}
\usepackage{adjustbox}
\newcommand{\dropcap}{}

\newcommand{\esm}[1]{\ensuremath{#1}}

\newcommand{\ms}[1]{\esm{\mathsf{#1}}}

\newcommand\xbase{x'}
\newcommand\reals{\mathbb{R}}
\newtheorem{definition}{Definition}
\newcommand\integratedgrads{\ms{IG}}

\newcommand\sparam{\alpha}
\newcommand{\synteq}{::=}

\newcommand{\stitle}[1]{\vspace{5pt} \noindent\textbf{#1.}\ }
\newcommand{\figthreewidth}{0.75}
\newcommand{\figonetwowidth}{0.8}
\newcommand{\phenylwidth}{0.2}
\newcommand{\pnasabstract}{}

\title{Using Attribution to Decode Dataset Bias in Neural Network Models for Chemistry}

\author{
  Kevin McCloskey \\
  Google Research\\
  1600 Amphitheatre Dr.\\
  Mountain View, CA \\
  \texttt{mccloskey@google.com} \\
  \And
  Ankur Taly \\
  Google Research\\
  1600 Amphitheatre Dr.\\
  Mountain View, CA \\
  \texttt{ankur.taly@gmail.com} \\
  \And
  Federico Monti \\
  Institute of Computational Science\\
  Università della Svizzera italiana\\
  Lugano, Switzerland
  \And
  Michael P. Brenner\\
  Google Research\\
  1600 Amphitheatre Dr.\\
  Mountain View, CA \\
  \And
  Lucy Colwell\\
  Google Research\\
  1600 Amphitheatre Dr.\\
  Mountain View, CA \\
  \texttt{lcolwell@google.com}\\
}

\begin{document}

\maketitle

\begin{abstract}
Deep neural networks have achieved state of the art accuracy at classifying molecules with respect to whether they bind to specific protein targets. A key breakthrough would occur if these models could reveal the fragment pharmacophores that are causally involved in binding. Extracting chemical details of binding from the networks could potentially lead to scientific discoveries about the mechanisms of drug actions. But doing so requires shining light into the black box that is the trained neural network model, a task that has proved difficult across many domains. Here we show how the binding mechanism learned by deep neural network models can be interrogated, using a recently described attribution method. We first work with carefully constructed synthetic datasets, in which the 'fragment logic' of binding is fully known. We find that networks that achieve perfect accuracy on held out test datasets still learn spurious correlations due to biases in the datasets, and we are able to exploit this non-robustness to construct adversarial examples that fool the model. The dataset bias makes these models unreliable for accurately revealing information about the mechanisms of protein-ligand binding. In light of our findings, we prescribe a test that checks for dataset bias given a  hypothesis. If the test fails, it indicates that either the model must be simplified or regularized and/or that the training dataset requires augmentation.
\end{abstract}

\pnasabstract

\dropcap{A} major stumbling block to modern drug discovery is to discover small molecules that bind selectively to a given protein target, while avoiding off-target interactions that are detrimental or toxic. The size of the small molecule search space is enormous, making it impossible to sort through all the possibilities, either experimentally or computationally \cite{polishchuk2013estimation}. The promise of {\sl in silico} screening is tantalizing, as it would allow compounds to be screened at greatly reduced cost \cite{shoichet2004virtual}. However, despite decades of computational effort to develop high resolution simulations and other approaches, we are still not able to rely solely upon virtual screening to explore the vast space of possible protein-ligand binding interactions \cite{schneider2017automating}. 

The development of high throughput methods for empirically screening large libraries of small molecules against proteins has opened up an approach where machine learning methods correlate the binding activity of small molecules with their molecular structure \cite{colwell2018statistical}. Among machine learning approaches, neural networks have demonstrated consistent gains relative to baseline models such as random forest and logistic regression \cite{dahl2014multi, ma2015deep, mayr2016deeptox, ramsundar2015massively, goh2017chemception}. In addition to protein-ligand binding, such models have been trained to predict physical properties that are calculated using density functional theory, such as polarizability and electron density \cite{MPNN_2017, schutt2017quantum, sinitskiy2018deep}. The ultimate promise of data-driven methods is to guide molecular design: models learned from ligands that bind to particular proteins will elucidate mechanism and generate new hypotheses of ligands that bind the required target in addition to improved understanding of the non-covalent interactions responsible.  

The motivating question for this work is: \emph{Why do virtual screening models make the predictions they do?} Despite their high accuracy, the major weakness of such data-driven approaches is the lack of causal understanding. While the model might correctly predict that a given molecule binds to a particular protein, it typically gives no indication of which molecular features were used to make this decision. Without knowledge of the binding logics the network can detect, or how the network architecture, training data and protocol play into this, it is not clear if the model learns the correct mechanism of binding, or spurious molecular features that happen to correlate with binding in the dataset being studied \cite{Atomwise_2017, brenner2016predicting, chuang2018adversarial}. Such model weaknesses are not captured by traditional evaluations based on measuring model accuracy over held out test sets because these datasets do not contain random samples drawn at uniform from the space of \emph{all} real-world molecules, but rather suffer from experimental selection bias. 

The key issue is to assess the influence of dataset biases on the prediction of protein binding. To unravel this, we establish a process for assessing whether a neural network model learns the correct logic of protein ligand binding, given a dataset of molecules with synthetic labels. Our method partitions molecules from the Zinc12 database \cite{Zinc_2012} according to specific binding logics, and then trains models on these sets of molecules and synthetic labels. We then use a recently developed attribution method \cite{IntegratedGradients_2017} to verify if each model learns the corresponding logic correctly. We develop a novel metric for attribution methods applied to models trained on synthetic labels, called the \emph{attribution AUC}, and use it to assess performance. 

The synthetic labels perfectly obey each binding logic, removing issues of experimental noise, so it is perhaps not surprising that we can train neural network models to perfectly predict the synthetic labels on held out test data in all cases. Nonetheless, the attribution AUC is often much lower than 1.0 (the perfect value), likely due to biases in the original dataset: since Zinc12 does not contain all possible molecules, there are molecular fragments that correlate with the binding logic yet are not themselves indicators of binding. This dataset bias implies that there exist ``adversarial molecules'' that do not satisfy the defined binding logic, for which the model makes incorrect predictions. Indeed, examining the model attributions allows us to identify adversarial molecules. Hence, even in this controlled setting, the network fails to learn the binding logic. Real-world protein-binding tasks are even more complex, due to noise in the binding assay, as well as underlying binding logics that are potentially more complex.

We apply this framework to ligands from the DUD-E dataset \cite{DUDE_2012} that bind ADRB2. We create a hypothesized logic for the binding mechanism, and create synthetic labels for the DUD-E dataset based on this logic.  Although a graph convolutional neural network makes perfect predictions on a held out dataset, biases in the dataset lead us to discover molecules which the model predicts bind to ADRB2, despite not satisfying the logic. The pattern used by the model to decide binding is different from the logic we imposed. Thus, despite its seemingly perfect performance, the model is fundamentally not able to predict that molecules bind for the right reason.

\section*{Analysis Framework}
\label{sec:methods}
\label{sec:dataset_construction}
To generate data with ground truth knowledge of the binding mechanism, we construct 16 synthetic binary label sets in which binding is \textit{defined} to correspond to the presence and/or absence of particular logical combinations of molecular fragments. For example, ligands could be labeled positive (i.e. bind to the target protein) if they obey the binding logic "carbonyl \textbf{and no} phenyl." Each binding logic is used to filter a large database of molecules to yield sets of positive and negative labeled molecules. In our implementation we specify molecular fragments using the SMARTS format \cite{SMARTS_2018} and we use RDKit \cite{RDKit_2017} to match them against candidate molecules, with a custom implementation of the logical operators \textbf{and}, \textbf{or}, and \textbf{not}. The 16 logics used in this paper are made up of elements sampled from 10 functional groups (Table S1), with up to four elements per logic joined by randomly selected operators (Tables 1, S2).

Dataset bias in chemistry is a well known issue that has previously been described~\cite{Atomwise_2017}. Essentially molecules that have been used in protein-ligand binding assays are not drawn uniformly at random from chemical space, but instead their selection for inclusion in a binding assay reflects the knowledge of expert chemists. These biases mean that large neural network models are at risk of overfitting to the training data. To reduce this risk, we carefully construct each dataset to be balanced, by sampling equally from all combinations of negations of the functional groups that make up each logic. In the case of just one functional group (A), this means that dataset contains equal numbers of molecules that match "A" and "\textasciitilde A". When there are two functional groups, say A and B, we have equal numbers matching "A\&B", "A\&\textasciitilde B", "\textasciitilde A\&B", and "\textasciitilde A\&\textasciitilde B". Similarly, all combinations are considered for logics with 3 and 4 functional groups. Each negation combination is represented by 1200 molecules in the dataset, with approximately 10\% of each reserved for held out model evaluation.

\begin{figure}
\centering
\includegraphics[width=\figonetwowidth\columnwidth]{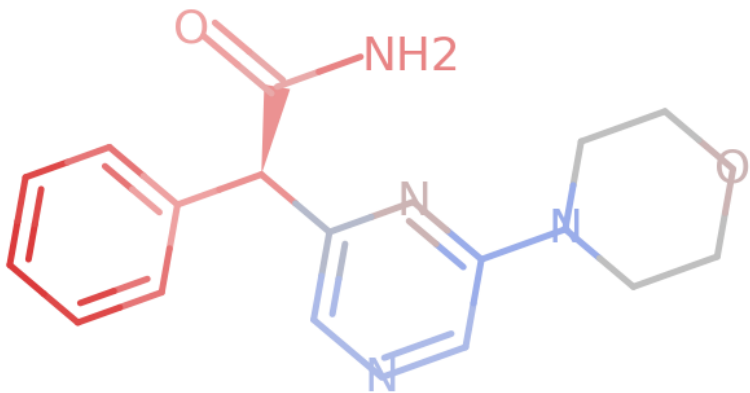}
\caption{An example of per-atom model attributions visualized for a molecule. Each atom is colored on scale from red to blue in proportion to its attribution score with red being the most positive and blue being the most negative.}
\label{fig:attribution_example}
\end{figure}

\stitle{Model Training}
With the datasets so defined, we then split the data into train and held out sets based on a hash of their Zinc ID, and train neural network models to predict 'binding' to these synthetic protein targets. We use two models: the molecular graph convolution (GC) model from Kearnes et al \cite{Kearnes_2016} and the message passing neural network (MPNN) from Gilmer et al \cite{MPNN_2017}. Both featurize each molecule using properties of atoms and pairs of atoms. We use the same hyperparameters reported in those papers, with the exception of a minibatch size of 99 and training each to only 10,000 steps. This took a little more than an hour on one GPU for each dataset. Given the lack of noise in the data, it is perhaps unsurprising that our models achieve nearly perfect cross-validation accuracies.

\stitle{Attribution Technique: Integrated Gradients}
We next seek to determine whether these models have learned the binding logic used to generate the synthetic labels. Given a trained model and an input, an attribution method assigns importance scores to each input feature in proportion to the contribution of that feature to the model prediction. Inspecting or visualizing the attribution scores reveals what features, in our case molecular fragments, were most relevant to the model's decision; see Figure~\ref{fig:attribution_example}. Formally, suppose a function $F: \reals^n \rightarrow [0,1]$ represents a deep network, which is given an input $x = (x_1,\ldots,x_n) \in \reals^n$. 
\begin{definition}\label{def:attribution}
An attribution of the prediction at input $x$ relative to a baseline input $\xbase$ is a vector $A_F(x, \xbase) =
(a_1,\ldots,a_n) \in \reals^n$ where $a_i$ is the \emph{contribution} of $x_i$ to the prediction $F(x)$. 
\end{definition}
In our case, the input $x$ is a molecule
featurized into properties of atoms and atom pairs, and $F(x)$ denotes the probability
of binding to a protein target. We normalize the attribution vector $A_F$ such that $\sum_i{a_i} = 1$. To compute attributions to individual molecular features we use a technique called ``Integrated Gradients''~\cite{IntegratedGradients_2017}. Attributions are defined relative to a baseline input, which essentially serves as the counterfactual in assessing the importance of each feature. Such counterfactuals are fundamental to causal explanations~\cite{KM86}. The typical baseline used for attribution on images is an image made of all black pixels; analogously here we use an input where all atom and atom-pair features are set to zero. See the Supplementary Information for more details.

The Integrated Gradient is defined as the path integral of the gradient along the linear
path from the baseline $\xbase$ to the input $x$. The intuition is as follows.
As we interpolate between the baseline and the input, the prediction moves 
along a trajectory, from uncertainty to certainty (the final probability).
At each point on this trajectory, one can use the gradient of the function $F$ with
respect to the input to attribute the change in probability back to the input variables.
The Integrated Gradient simply aggregates this gradient along this trajectory using a path integral.
\begin{definition}[\textbf{Integrated Gradients}]\label{def:intgrad}
Given an input $x$ and baseline
$\xbase$, the integrated gradient along the $i^{th}$ dimension is defined as follows.
\begin{equation}\label{eqn:intgrad}
\integratedgrads_i(x,\xbase) \synteq (x_i-\xbase_i)\times\int_{\sparam=0}^{1} \tfrac{\partial F(\xbase + \sparam\times(x-\xbase))}{\partial x_i  }~d\sparam
\end{equation}
(here
$\tfrac{\partial F(x)}{\partial x_i}$ is the gradient of
$F$ along the $i^{th}$ dimension at $x$).
\end{definition}

The attributions computed in this way assign scores to both atom and atom-pair features. To simplify
the analysis, we distribute the atom pair scores evenly between the individual atoms present in each pair. Formally, if $a_i$ is the attribution computed for atom $i$, and $e_{ij}$ is the attribution computed for atom pair $i,j$, then our 
aggregated attribution score for atom $i$ is
\begin{equation}
\tilde{a}_i = a_i + \sum_{j \in E_i}{\frac{e_j}{2}} 
\end{equation}
where $E_i$ is the set of all featurized atom-pairs that include atom $i$. Henceforth we only study these aggregated per-atom attributions for each molecule.

There are several other methods for attributing a deep network's prediction to its input
features. We selected Integrated Gradients due to its ease of implementation, wide applicability, 
and demonstrated effectiveness over a variety of network architectures and tasks.
It is also justified by an axiomatic result, that is, 
it is essentially the unique method that satisfies certain desirable properties 
of an attribution method. We refer to~\cite{IntegratedGradients_2017} for formal definitions, results, and a thorough comparison to alternate attribution methods.

\begin{figure}
  \centering
  \includegraphics[width=\figonetwowidth\columnwidth]{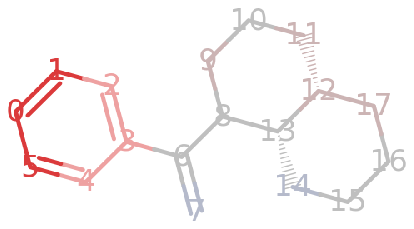}
  
  \begin{tabular}{p{0.3\columnwidth}p{0.3\columnwidth}p{0.3\columnwidth}}
    Atom index & \parbox{0.3\columnwidth}{Attribution Score \\Ranking} & \parbox{0.3\columnwidth}{Label \\for AUC \\calculation} \\
    \midrule
    1 & 0.29 & 1 \\
    5 & 0.29 & 1 \\
    0 & 0.28 & 1 \\
    2 & 0.09 & 1 \\
    4 & 0.09 & 1 \\
    3 & 0.07 & 1 \\
    9 & 0.03 & 0 \\
    11 & 0.02 & 0 \\
    ... & ... & ... \\
    \bottomrule
  \end{tabular}
  
  \caption{Top, visualization of Integrated Gradients on a "binding" molecule from the Logic 1 dataset that requires a phenyl group. Bottom, the top 8 atoms seen above as ranked by the attribution score. This molecule would receive an Attribution AUC of 1.0 for these attributions.}
  \label{fig:attribution_auc}
\end{figure}

\stitle{Attribution AUC}
Ideally, we would like the attribution scores to isolate the synthetic binding logic used to label the dataset, since this would translate to the ability to identify pharmacaphores in real data. Attribution scores are typically studied by visualization using heatmaps; figure~\ref{fig:attribution_example} provides a visualization of the per-atom attribution scores for a molecule. If a model learns the correct binding logic, we would expect the attribution scores to be large in magnitude for atoms involved in the binding logic and small elsewhere. To measure this quantitatively, we propose a metric, called the \emph{attribution-AUC}, that compares attribution scores against a given binding logic. We calculate the attribution-AUC score for each molecule separately and report the average across all molecules in the held out set for each dataset. (The attribution-AUC is entirely distinct from the \emph{model AUC} which measures the overall accuracy of the model on the binding task.)

Figure \ref{fig:attribution_auc} illustrates the attributions calculated for a molecule using the model trained on synthetic labels. A positive attribution score (red) indicates that this atom increases "protein binding" ability, according to the trained model, whereas a negative attribution score (blue) indicates that the model thinks that this atom hurts binding. Our goal is to evaluate how faithfully these scores reflect the binding logic used to label the dataset. 

There are two ways in which an atom can participate in a binding logic - as part of a fragment that is required to be either present or absent for binding to take place. Our attribution-AUC calculation handles these cases separately. For those fragments required to be present, we assign each fragment atom the label 1, and all other atoms the label 0. We then use these labels and the attribution scores in SKLearn's \cite{scikit-learn} ROC AUC implementation to compute the present-attribution-AUC. The process for those fragments required to be absent is exactly analogous, except that we first multiply all attribution scores by -1.0 before calculating the absent-attribution-AUC. The final attribution-AUC for the molecule in question is simply the average of the present-attribution-AUC and the absent-attribution-AUC. This same process is applied regardless of which synthetic "binding" label the molecule received during dataset generation.

For more complex logics, the AUC calculation is more subtle. With disjunctive logics, that require the presence of either fragment A, or fragment B, or both fragments A and B, we calculate the attribution-AUC using the atom labelings for all possible disjunctive fragment combinations and report the maximum attribution-AUC found for each molecule. Similarly, for molecules which exhibit multiple instances of the same functional group or fragment, it is difficult to decide on a single correct labeling for the attribution. In this case we similarly calculate the attribution-AUC for the disjunctive combination of all fragment instances present in each molecule. We report the maximum attribution-AUC found using this approach for each molecule in the held out test set.

\begin{table*}[!h]
\centering
\begin{adjustbox}{width=1\textwidth}
\begin{tabular}{ccccc}
  Synthetic binding logic & \parbox{0.1\columnwidth}{\centering GC \\Model \\AUC} & \parbox{0.1\columnwidth}{\center GC \\Attribution \\AUC} & \parbox{0.1\columnwidth}{\centering MPNN \\Model \\AUC} & \parbox{0.1\columnwidth}{\centering MPNN \\Attribution \\AUC} \\
  \midrule
  \mbox{unbranched alkene} & 1.0 & 0.98 & 0.99 & 0.99 \\
  \mbox{phenyl} & 0.995 & 0.98 & 1.0 & 0.99 \\
  \mbox{napthalene} & 1.0 & 1.0 & 1.0 & 1.0  \\
  \mbox{\textbf{no} primary amine} & 1.0 & 0.97 & 1.0 & 1.0 \\
  \mbox{ether \textbf{or} \textbf{no} alkyne} & 0.992 & 0.91  & 1.0 & 0.9\\
  \mbox{naphthalene \textbf{and} \textbf{no} primary amine} & 0.999 & 0.89  & 1.0 & 0.77\\
  \mbox{fluoride \textbf{and} carbonyl} & 1.0 & 0.77 & 1.0 & 0.61 \\
  \mbox{unbranched alkane \textbf{and} carbonyl} & 1.0 & 0.79  & 1.0 & 0.83\\
  \mbox{fluoride \textbf{and} alcohol \textbf{and} \textbf{no} alkene} & 1.0 & 0.93 & 0.99 & 0.9 \\
  \mbox{primary amine \textbf{and} ether \textbf{and} phenyl} & 0.995 & 0.7 & 0.99 & 0.6 \\
  \mbox{\textbf{(}primary amine \textbf{or} \textbf{no} phenyl\textbf{)} \textbf{and} \textbf{no} ether} & 0.999 & 0.86 & 1.0 & 0.83 \\
  \mbox{alcohol \textbf{and} \textbf{no} fluoride \textbf{and} \textbf{no} ether} & 1.0 & 0.88 & 1.0 & 0.85 \\
  \mbox{floride \textbf{and} unbranched alkane \textbf{and} \textbf{(}primary amine \textbf{or} \textbf{no} alcohol\textbf{)}} & 0.999 & 0.67 & 0.94 & 0.66 \\
  \mbox{\textbf{(}phenyl \textbf{and} \textbf{no} carbonyl\textbf{)} \textbf{or} \textbf{(}alkyne \textbf{and} \textbf{no} ether\textbf{)}} & 1.0 & 0.7 & 1.0 & 0.67 \\
  \mbox{\textbf{(}ether \textbf{or} \textbf{no} alcohol\textbf{)} \textbf{and} carbonyl \textbf{and} \textbf{no} alkyne} & 1.0 & 0.75  & 1.0 & 0.71\\
  \mbox{unbranched alkane \textbf{and} \textbf{no} ether \textbf{and} primary amine \textbf{and} carbonyl} & 0.996 & 0.76 & 0.98 & 0.62\\
\bottomrule
\end{tabular}
\end{adjustbox}
\caption{This table shows the attribution-AUC (computed using Integrated Gradients) and the model AUC for Graph Convolution networks and MPNNs trained against synthetic data labels generated according to the binding logics listed in column 1. See the Supplementary Information for more details on the binding logics and their component molecular fragments.}
\label{tab:main_results}
\end{table*}

\section*{Results}
Table~\ref{tab:main_results} lists both the model AUC and the attribution-AUC results obtained for networks trained using data with synthetic labels that reflect the binding logics listed. The model AUC is near perfect (1.0) for each of the binding logics indicating that the trained models can correctly classify the molecules in the held-out test sets. However the attribution-AUC is significantly lower than 1.0 for several logics. For instance, for binding logic 9, which can be read as ``primary amine and ether and phenyl.'', the GC attribution-AUC is only $0.7$ while the model AUC is $0.995$. Furthermore, we note that the attribution-AUC scores decline as the logics become more complicated and include larger numbers of functional groups. The MPNN results also exhibits a similar pattern. We now discuss the implications of these findings.

\stitle{Attacks guided by attributions}\label{sec:attacks}
The combination of near-perfect model performance and low attribution-AUCs revealed in Table~\ref{tab:main_results} indicate either: (1) a weakness of the attribution technique, or (2) a weakness of each model at correctly learning the synthetic binding logics. We distinguish these cases by investigating individual examples of molecules that were correctly classified but whose attribution-AUCs were low. Guided by patterns across multiple molecules where the attributions were misplaced with respect to the ground truth described by the binding logic, we discovered small perturbations of each molecule which caused the class predicted by the model to be incorrect. We were able to find at least one perturbation attack for every logic that did not have a high attribution-AUC, leading us to conclude that the model did not learn the correct binding logic in these cases. These results also clarify the fact that the heldout sets are still under-representative, despite their careful balancing, discussed above.

\begin{figure}
\includegraphics[width=\figthreewidth\columnwidth]{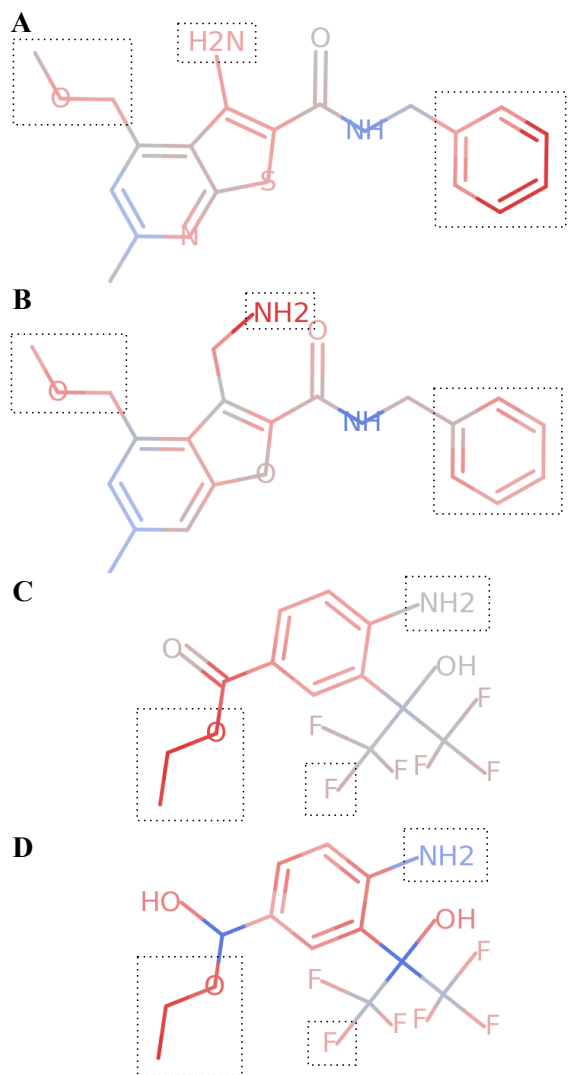}


  \caption{Visualizations of attribution scores, calculated using Integrated Gradients. A) Attribution scores for a molecule from the logic 9 heldout set that obeys the binding logic. B) A minor perturbation of the above molecule, guided by errors in the attributions shown in (A), which gets misclassified by the model. C) Attribution scores for a molecules from the logic 12 heldout set that obeys the binding logic. D) A minor perturbation of the above molecule which still obeys the logic, but is misclassified by the model. Dotted boxes are added around the fragments whose presence defines the molecules as members of the positive class.}
\label{fig:attack}
\end{figure}

Here, we describe a few of the perturbation attacks that we found. Binding logic 9 requires the presence of "a primary amine and an ether and a phenyl." One example from Zinc that satisfies this logic is shown in Figure~\ref{fig:attack}A. This molecule is correctly classified as positive (i.e. binding) by the model with a probability of 0.97, however as seen in the figure it has misplaced attributions on several atoms in the ring structures on the left. We perturb those atoms and separate the primary amine from them with an additional carbon, resulting in the molecule shown in Figure~\ref{fig:attack}B. The model gives this perturbed molecule a predicted score of 0.20, a negative class prediction, despite the fact that the molecule still fully satisfies the same binding logic that the model was trained against.

Binding logic 12 requires that a molecule satisfy the "absence of an alcohol or presence of a primary amine, along with an unbranching alkane and a fluoride group." One example from Zinc that satisfies this logic is shown in Figure~\ref{fig:attack}C. It is correctly classified as positive by the model with a prediction of 0.97, however it has misplaced attributions on the carbon atom in the carbonyl group on the left. Guided by these attributions we perturb that carbonyl, converting it to a single bond, resulting in the molecule in Figure~\ref{fig:attack}D. The model gives this perturbed molecule a predicted score of 0.018, a negative class prediction, despite the fact that the molecule still satisfies the ground truth binding logic.

\stitle{A pharmacological hypothesis}
These results indicate that the attribution can be more trustworthy than the model: even if the model achieves a high AUC, a low attribution-AUC appears to indicate that there exist molecules that do not satisfy the binding logic but are predicted to bind by the model.  This occurs because of biases in the underlying dataset, leading the model to confound binding with dataset bias. 

The same concern applies to real protein binding datasets. Our results suggest a simple test that can be performed to test an existing hypothesis about the pharmacophore(s) that control binding. First, the  hypothesis is codified as a ``binding logic'', which is used to create a set of synthetic labels.  Next, these synthetic labels are used to train a neural network and analyze its attributions and attribution-AUC. A high AUC, with attribution to the correct functional groups suggests that the combination of dataset and trained neural network is able to generalize. However, a low attribution-AUC or consistent unexpected attribution artifacts would suggest a need for model simplification and regularization, and/or dataset augmentation.

\begin{figure}

  \includegraphics[width=0.9\columnwidth]{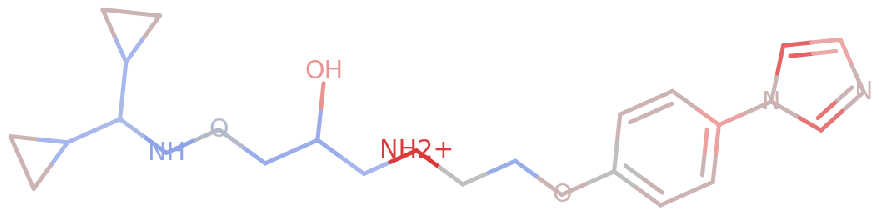}
  \includegraphics[width=0.9\columnwidth]{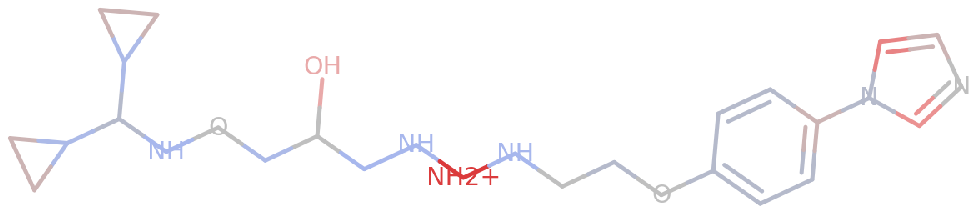}

  \caption{Visualizations of Integrated Gradients attributions. Top, on an example "binder" from the synthetic ADRB2 dataset, correctly predicted as a positive with prediction 0.999. Bottom, a minor perturbation of the above molecule which should be a negative but gets misclassified as still a positive with prediction 0.995.}
  \label{fig:attack_adrb2}
\end{figure}

We follow this protocol using data for binding to the protein ADRB2 from the DUD-E dataset \cite{DUDE_2012}. One hypothesis for a pharmacaphore is a benzene ring with a two-carbon chain connected to an ionized secondary amine, i.e. \{[cR1]1[cR1][cR1][cR1][cR1][cR1]1\}\&\{CC[NH2+]\}. This results in a dataset with 934 positives and 14290 negatives, of which \textasciitilde10\% are reserved as a heldout set by ID hash. We trained a graph convolution model (see details in SI text), and achieved a Model AUC on the heldout set of 1.0.  However its attribution-AUC is extremely low, at only 0.11. Visualizations of the attributions show the attribution only consistently highlights the NH2+ group. This means that attacks (e.g.  Figure~\ref{fig:attack_adrb2}) are easily discovered using this insight.

This analysis suggests a weakness in the ADRB2 dataset and trained neural network. Our findings are aligned with the conclusions of \cite{Atomwise_2017} around dataset bias, but notably our proposal considers not just the data but the role of the neural network itself: we expect that different architectures and different hyperparameters will exhibit overfitting of the same dataset differently.

\section*{Discussion}\label{sec:discussion}
There is growing concern about the non-robustness of machine learning models, and much recent research has been devoted to finding ways to assess and improve model robustness~\cite{MTSD18,Papernot-thesis18,RSG18,ZAG18,ZDS18,Dixon18,Atomwise_2017, brenner2016predicting, chuang2018adversarial}. A common source of non-robustness is bias in the training dataset~\cite{Atomwise_2017, MTSD18,RSG18,Dixon18}. An approach to identifying such bias is to examine attributions of the model's predictions, and determine if too much attribution falls on non-causal features or too little falls on causal features~\cite{MTSD18}; both are undesirable and indicative of bias in the training dataset.

The central challenge in applying this approach to virtual screening models is that 
 {\sl a priori}, we know neither the internal logic of the model, nor the logic of protein binding. Thus we have no reference for assessing the attributions. To resolve this, we introduce the idea of evaluating hypotheses for binding logics by setting up a synthetic machine learning task. We use the hypothesized logic to relabel molecules used in the original study, and train a model to predict these labels. If attributions fail to isolate the hypothesized logic on this synthetic problem, it signals that there exist biases in the training data set that fool the model into learning the wrong logic. Such bias would also likely affect the model's behavior on the original task.

To quantitatively assess attributions, we introduce the attribution-AUC metric, measuring how well the attributions isolate a given binding logic. It is not a measure of the ``correctness'' of the attributions. Recall that the mandate for an attribution method is to be faithful to the model's behavior, and not the behavior expected by the human analyst~\cite{IntegratedGradients_2017}. In this work, we take the faithfulness of the attributions obtained using Integrated Gradients as a given.
Indeed, for our synthetic task, we find the attributions to be very useful in identifying biases in the model's behavior, and we were able to successfully translate such biases into perturbation attacks against the model. These attacks perturb those bonds and atoms with unexpected attributions, and their success also confirms the faithfulness of the attributions. The attacks expose flaws
in the model's behavior despite the model having perfect accuracy on a held out test set. This reiterates the risk of solely relying on held out test sets to assess model behavior.

An important caveat here is that realistic attacks require the molecule perturbations to respect the laws of chemistry. For instance, simply dropping an atom may create a molecule that is invalid per the rules of valency. The model's prediction may be affected not because of the absence of the atom but because of the artifact that is introduced. Thus, the role of the human analyst is key in ensuring that the perturbations are plausible. Finally, we acknowledge that attributions as a tool offer a very reductive view of the internal logic of the model. They are analogous to a first-order approximation of a complex non-linear function. They fail to capture higher order effects such as how various input features interact during the computation of the model's prediction. Such interactions between atom and bond features are certainly at play in virtual screening models. Further research must be carried out to reveal such feature interactions.

\stitle{Thoughts for practitioners}
The recent machine learning revolution has led to great excitement regarding the use of neural networks in chemistry. Given a large dataset of molecules and quantitative measurements of their properties, a  neural network can learn/regress the relationship between features of the molecules and their measured properties. The resulting model can have the power to predict properties of molecules in a held out test set, and indeed can be used to find other molecules with these properties. Despite this promise, an abundance of caution is warranted: it is dangerous to trust a model whose predictions one does not understand. A serious issue with neural networks is that although a held out test set may suggest that the model has learned to predict perfectly, there is no guarantee that the predictions are made for the right reason. Biases in the training set can easily cause errors in the model's logic. The solution to this conundrum is to take the model seriously: analyze it, ask it why it makes the predictions that it does, and avoid relying solely on aggregate accuracy metrics. The attribution-guided approach described in this paper for evaluating learning of hypothesized binding logics may provide a useful starting point.

\stitle{Acknowledgements}
We would like to thank Steven Kearnes and Mukund Sundararajan for helpful conversations. MPB gratefully acknowledges support from the National Science Foundation through NSF-DMS1715477, as well as support from the Simons Foundation. FM is supported in part by ERC Consolidator Grant No. 724228 (LEMAN), Google Faculty Research Awards, an Amazon AWS Machine Learning Research grant, an Nvidia equipment grant and a Rudolf Diesel Industrial Fellowship at IAS TU Munich.

\bibliography{Arxiv-main}{}
\bibliographystyle{plain}

\appendix
\onecolumn
\raggedbottom
\renewcommand{\thetable}{S\arabic{table}}
\setcounter{table}{0}

\begin{center}
\begin{Huge}
Supplementary Information
\end{Huge}
\end{center}

\vspace{10pt}

\subsection*{Integrated Gradients Baseline}
\label{sec:null_molecules}
The baseline molecule is an important part of the Integrated Gradients technique. 
Specifically, the technique computes attributions for a given input relative to a certain baseline
input. Furthermore, it can be shown that attributions to all features of the input
sum up to the difference between the model prediction for the input and the 
model prediction for the baseline input. 
\[ \sum_i{ \int_{\alpha=0}^{1}{ \frac{\partial F(x\cdot\alpha)}{\partial(x_i\cdot\alpha)} \frac{\partial(x_i\cdot\alpha)}{\partial \alpha}}\partial \alpha} \]
\[ = \int_{\alpha=0}^{1}{\bigg( \frac{\partial F(x\cdot\alpha)}{\partial(x\cdot\alpha)} \bigg)^T \frac{\partial(x\cdot\alpha)}{\partial \alpha} \partial \alpha } \]
\[ = F(x\cdot1) - F(x\cdot0)\]

\noindent Ideally, the baseline input must be an uninformative input with a neutral prediction. In our case, we set the baseline to a molecule whose atom and atom-pair features are zero. In order for the model to have a neutral prediction for this baseline molecule, we found it necessary to augment the training data with examples of the baseline molecule. We did this by splitting each minibatch of 99 examples as follows: 33 examples are drawn at random from the Zinc-filtered and balanced dataset for each logic (which by design contains equal numbers of positive and negative examples), 33 were baseline molecules labeled as 1.0, and 33 were baseline  molecules labeled as 0.0. This achieves the desired result that the baseline molecule receives a model prediction of 0.5. This score of 0.5 is the ideal neutral point because it is far from both positive-class final predictions ($\approx 1.0$) and negative-class final predictions ($\approx 0.0$). 
Finally, we note that in this work we discretized the integration path with 90 step values of the $\alpha$ parameter.

\subsection*{ADRB2 Model Training}
When training on the synthetic ADRB2 dataset, we used the GC model and training settings as described in the section Analysis Framework, \textit{Model Training}, with one exception: to balance the training data we used a minibatch size of 100, and split each minibatch equally 4 ways: 25 molecules were positive examples of ``binders'', 25 molecules were negative examples of ``nonbinders'', 25 molecules were 'baseline' molecules labeled 1.0, and 25 were 'baseline' molecules labeled 0.0.

\subsection*{Training data from ZINC}
The compounds in the 16 synthetic logic datasets we analyze in this work are drawn from ZINC12 (https://zinc.docking.org). If you are interested in reproducing or comparing to this work, we provide these datasets in an ancillary file available for download from the Arxiv (https://arxiv.org/abs/1811.11310). This is a static copy of the data. For the latest purchasing information, it is best to visit the latest version (https://zinc15.docking.org/). We are distributing this subset of ZINC with John Irwin's permission.

\begin{table*}
\centering
\begin{tabular}{l|l|c}
Functional group & SMARTS & Visualization \\
\hline\hline
alkene & [CX3]=[CX3] & \includegraphics[width=0.25\columnwidth]{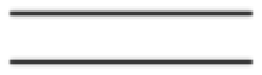} \\ \hline
unbranched alkane & [R0;D2,D1][R0;D2][R0;D2,D1] & \includegraphics[width=0.25\columnwidth, height=15mm]{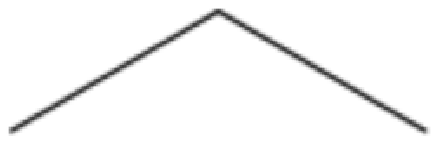} \\ \hline
alkyne & [CX2]\#[CX2] & \includegraphics[width=0.25\columnwidth, height=15mm]{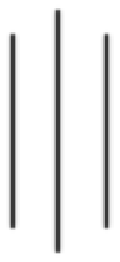} \\ \hline
primary amine & [NX3;H2] & \includegraphics[width=0.25\columnwidth]{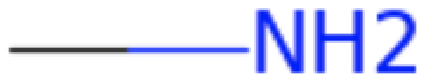} \\ \hline
alcohol & [OX2H] & \includegraphics[width=0.25\columnwidth]{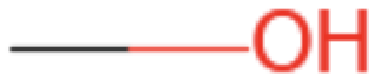} \\ \hline
ether & [OD2](C)C & \includegraphics[width=0.25\columnwidth]{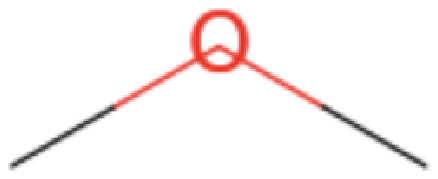} \\ \hline
carbonyl & [CX3]=O & \includegraphics[width=0.25\columnwidth]{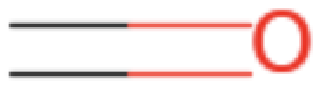} \\ \hline
fluoride & [FX1] & \includegraphics[width=0.25\columnwidth]{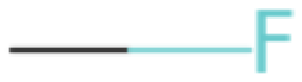} \\ \hline
naphthalene & c1:c:c:c2:c:c:c:c:c:2:c:1 & \includegraphics[width=0.25\columnwidth]{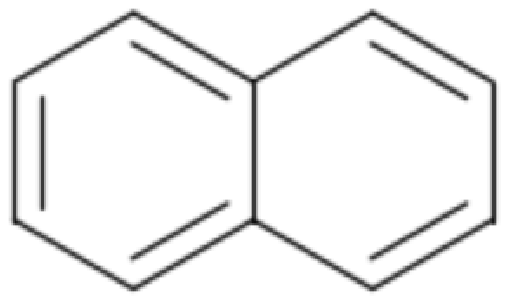} \\ \hline
phenyl & [cX3]1[cX3H][cX3H][cX3H][cX3H][cX3H]1 & \includegraphics[width=\phenylwidth\columnwidth, height=25mm]{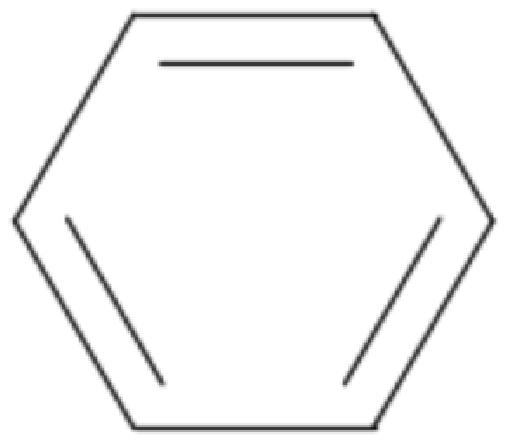} \\ 
\hline\hline
\end{tabular}
\caption{The organic chemistry functional groups used in the creation of the synthetic datasets.}
\label{tab:functional_groups}
\end{table*}

\begin{table*}
\centering
\begin{adjustbox}{width=1\textwidth}
\begin{tabular}{ll}
Logic \# & Logic \\
\midrule
0 & [R0;D2,D1][R0;D2][R0;D2,D1] \\
1 & [cX3]1[cX3H][cX3H][cX3H][cX3H][cX3H]1 \\
2 & c1:c:c:c2:c:c:c:c:c:2:c:1 \\
3 & \textasciitilde\{[NX3;H2]\} \\
4 & \{\textasciitilde\{[CX2]\#[CX2]\}\}|\{[OD2](C)C\} \\
5 & \{c1:c:c:c2:c:c:c:c:c:2:c:1\}\&\{\textasciitilde\{[NX3;H2]\}\} \\
6 & \{[FX1]\}\&\{[CX3]=O\} \\
7 & \{[R0;D2,D1][R0;D2][R0;D2,D1]\}\&\{[CX3]=O\} \\
8 & \{\{\textasciitilde\{[CX3]=[CX3]\}\}\&\{[OX2H]\}\}\&\{[FX1]\} \\
9 & \{\{[NX3;H2]\}\&\{[OD2](C)C\}\}\&\{[cX3]1[cX3H][cX3H][cX3H][cX3H][cX3H]1\} \\
10 & \{\{\textasciitilde\{[cX3]1[cX3H][cX3H][cX3H][cX3H][cX3H]1\}\}|\{[NX3;H2]\}\}\&\{\textasciitilde\{[OD2](C)C\}\} \\
11 & \{\{\textasciitilde\{[FX1]\}\}\&\{\textasciitilde\{[OD2](C)C\}\}\}\&\{[OX2H]\} \\
12 & \{\{\{\textasciitilde\{[OX2H]\}\}|\{[NX3;H2]\}\}\&\{[R0;D2,D1][R0;D2][R0;D2,D1]\}\}\&\{[FX1]\} \\
13 & \{\{\{[cX3]1[cX3H][cX3H][cX3H][cX3H][cX3H]1\}\&\{\textasciitilde\{[CX3]=O\}\}\}|\{[CX2]]\#[CX2]\}\}\&\{\textasciitilde\{[OD2](C)C\}\} \\
14 & \{\{\{\textasciitilde\{[OX2H]\}\}|\{[OD2](C)C\}\}\&\{[CX3]=O\}\}\&\{\textasciitilde\{[CX2]\#[CX2]\}\} \\
15 & \{\{\{\textasciitilde\{[OD2](C)C\}\}\&\{[R0;D2,D1][R0;D2][R0;D2,D1]\}\}\&\{[NX3;H2]\}\}\&\{[CX3]=O\} \\
\bottomrule
\end{tabular}
\end{adjustbox}
\caption{The 16 logics we studied. Molecular fragments are in SMARTS format combined with
\textit{and} (\{...\}\&\{...\}), \textit{or} (\{...\}|\{...\}) and \textit{not} (\textasciitilde\{...\}) operators. }
\label{tab:logics}
\end{table*}

\end{document}